\documentclass[11pt]{article}

\usepackage[final]{acl}

\usepackage{times}
\usepackage{latexsym}
\usepackage{amsmath}
\usepackage{enumitem}
\usepackage{times}
\usepackage{latexsym}
\usepackage{hyperref}
\usepackage{graphicx}
\usepackage{xcolor}
\usepackage[dvipsnames]{xcolor}
\usepackage{amsmath}
\usepackage{booktabs}
\usepackage{enumitem}
\usepackage{bbding}
\usepackage{longtable}
\usepackage{pdflscape}
\usepackage{afterpage}
\usepackage{adjustbox}
\usepackage{makecell}
\usepackage{booktabs}
\usepackage{comment}
\usepackage{arydshln}
\usepackage{amssymb}
\usepackage{bm}
\usepackage{subcaption}
\usepackage{caption}
\usepackage{textcomp}
\usepackage{fixltx2e}
\usepackage{float}
\usepackage{array}
\usepackage{rotating}
\usepackage{tabularx}
\usepackage{multirow}

\usepackage[T1]{fontenc}

\usepackage[utf8]{inputenc}

\usepackage{microtype}

\usepackage{inconsolata}

\usepackage{graphicx}

\title{Can We Optimize the Performance-Carbon Emission Break-Even Point?: The Quest for Greener LLMs}

\author{
Sourav Das\textsuperscript{*} \\
IIIT Kalyani \\
\texttt{\small{sourav\_phd21@iiitkalyani.ac.in}} \\\And
Tanmay Joshi\textsuperscript{*} \\
BITS Pilani Goa \\
\texttt{\small{f20231102@goa.bits-pilani.ac.in}} \\\And
Kripabandhu Ghosh \\
IISER Kolkata \\
\texttt{\small{kripa.ghosh@gmail.com}} }

\begin{document}
\maketitle
\begin{abstract}
The carbon footprint of any deployed Large Language Model (LLM) accumulates during inference, where repeated use of the model substantially exceeds the one-time cost of fine-tuning. Yet most efficiency interventions target either pre-training scale or post-hoc compression. We ask whether folding a calibrated, differentiable energy surrogate into the fine-tuning objective can produce inference behavior that gains task accuracy at zero or near-zero carbon cost, a break-even configuration. We propose a joint loss mechanism with a per-model carbon-emission parameter, a linear surrogate over parameter norm, FLOP proxy, and memory proxy, fit from on-hardware energy profiling. We fine-tune three architecturally distinct families: Gemma-2 2B, Llama-3.1 8B, and Qwen-2.5 14B, and evaluate inference F1 and CO$_2$ emissions on three MMLU subjects: abstract algebra, philosophy, and formal logic. We discover from several outcomes that the carbon term behaves as either harmful interference or beneficial regularization depending on the task structure. We position calibrated carbon-aware fine-tuning as a lightweight, drop-in regularizer with a non-empty but model and task-dependent break-even region. \textbf{This is an ongoing work, and we will release our codebase soon.}
\end{abstract}


\begingroup
\renewcommand{\thefootnote}{}
\footnotetext{*These authors contributed equally.}
\addtocounter{footnote}{-1}
\endgroup

\section{Introduction}
\label{sec:intro}

\begin{figure}[t]
\centering

\includegraphics[width=\columnwidth]{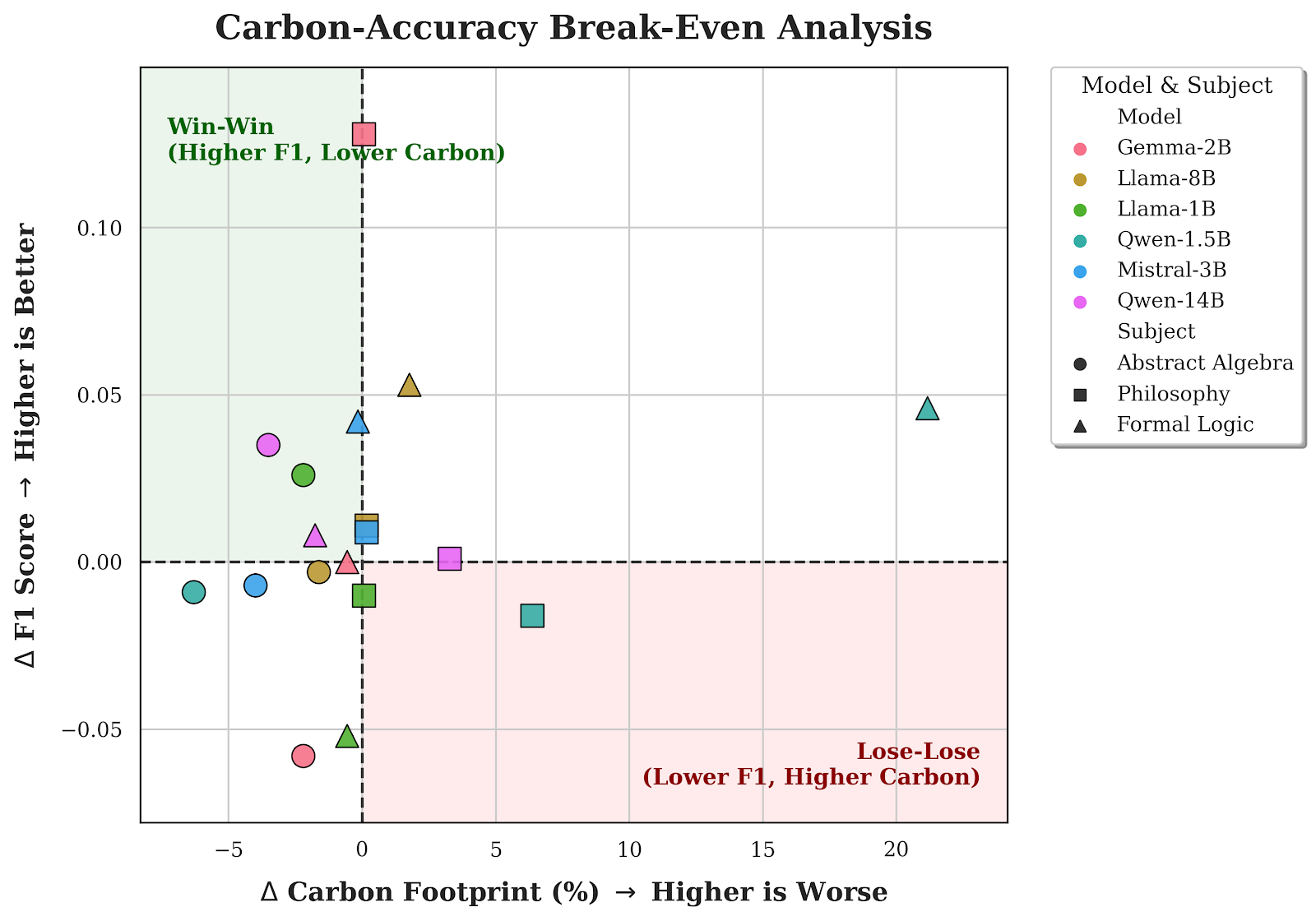}
\caption{\textbf{The carbon--accuracy break-even region.}
Each marker is one (model family, MMLU subject) pair, showing
inference $\Delta$F1 and relative $\Delta$CO$_2$ of our joint-loss
model against a cross-entropy baseline. Points in the upper-left
quadrant are strict Pareto improvements; the dashed line marks
zero carbon delta. Five of nine pairs lie inside the break-even
region ($\Delta$F1$\,\geq\,0$ and $\Delta$CO$_2\,\leq\,+2\%$),
including one strict Pareto improvement on Qwen-14B / abstract
algebra (Section \ref{sec:results}).}
\label{fig:teaser}
\end{figure}

The carbon footprint of LLMs accumulates over its operational lifetime rather than at the moment of training. A frontier LLM is fine-tuned a handful of times but serves significantly more forward passes when in production. Recent measurement and accounting studies argue that inference, not training, governs the long-run environmental cost of widely used systems \citep{patterson2021carbon, wu2022sustainable, luccioni2024power}. The community's response has nevertheless concentrated at the two ends of the lifecycle. Training-time work pursues better scaling laws and parameter-efficient adaptation \citep{hoffmann2022chinchilla, hu2022lora, dettmers2023qlora}; post-training work targets the frozen artifact through quantization, pruning, distillation, and decoding-time acceleration \citep{frantar2023gptq, frantar2023sparsegpt, xiao2023smoothquant, leviathan2023speculative, sun2024wanda}. The fine-tuning objective itself, the mechanism that fixes which computational pathways the deployed model will exercise for every subsequent query, has remained, in effect, carbon-neutral terrain.

We argue this is a missed leverage point. The loss function chosen during fine-tuning quietly determines the model's inference-time circuit usage, and a calibrated energy term inserted at this stage can steer the model toward lower-cost pathways without modifying its architecture, its precision, or its decoding routine. Differentiable hardware-aware neural architecture search established this principle in the vision domain a half-decade ago, embedding FLOP and latency proxies into the search objective and obtaining accurate, low-energy models with no post-hoc compression step \citep{wu2019fbnet, cai2019proxylessnas, tan2019mnasnet}. The corresponding move for LLMs folding a differentiable, on-hardware energy surrogate into the fine-tuning loss itself has not been studied, to the best of our knowledge. Existing carbon-aware LLM work measures and reports emissions \citep{strubell2019energy, schwartz2020green, luccioni2023estimating}, but does not optimize against them at training time.

Our paper asks a single prominent research question: \emph{Can joint optimization of task performance and inference carbon emission reach a break-even point?} By \emph{break-even} we mean an operating point at which downstream task F1 is preserved or improved while inference CO$_2$ is no higher than that of a standard cross-entropy baseline.We fine-tune three architecturally distinct families: Gemma-2-2B \citep{gemmateam2024gemma2}, Llama-3.1-8B \citep{dubey2024llama3}, and Qwen-2.5-14B \citep{qwen2025qwen25}, and evaluate inference F1 and CO$_2$ on three MMLU subjects \citep{hendrycks2021mmlu}: abstract algebra, philosophy, and formal logic.

Figure~\ref{fig:teaser} represents that the break-even region is non-empty, though selective. Qwen-14B on abstract algebra delivers a strict Pareto improvement, gaining $3.5$ F1 points while reducing inference CO$_2$ by $3.5\%$. Gemma-2B on philosophy gains $12.8$ F1 points at essentially zero carbon delta, and Llama-8B on formal logic gains $5.3$ F1 points for a $1.8\%$ carbon increase, a regime in which the per-query F1 yield substantially exceeds the marginal emissions cost. Taken together, these results suggest that carbon-aware fine-tuning is best understood not as a uniform efficiency intervention but as a structural regularizer whose effect is mediated by the magnitude of the target task. \textit{To our knowledge, no prior work attempted a differentiable energy surrogate of a set of pretrained LLMs into the fine-tuning loss to navigate the trade-off between task accuracy and inference CO$_2$ emissions toward a break-even configuration.}

We make four novel contributions in this work, two primary and two subsidiary:
\begin{enumerate}[nosep]
    \item We introduce \emph{calibrated carbon-aware fine-tuning} for LLMs: a drop-in joint loss that adds a differentiable energy surrogate, fitted from on-hardware profiling of the specific model under training, directly to the task objective (Section \ref{sec:method:loss}, Section \ref{sec:method:surrogate}).
    \item Across three model families and three MMLU subjects, we identify a non-empty break-even region containing one strict Pareto improvement, demonstrating that the regime is reachable in practice (Section \ref{sec:results:mmlu}).
    \item Through a $\lambda$-sensitivity study on Qwen-14B over SQuAD and BoolQ, we show that the optimal carbon penalty is task-structure-conditional, reframing $\lambda$ as a regularizer whose interpretation shifts between tasks (Appendix \ref{sec:results:lambda}).
    \item We release per-step training histories, calibration tables, inference-time emissions logs, and MMLU prediction files for all configurations, enabling downstream replication (Appendix \ref{app:reproducibility}).
\end{enumerate}

\section{Method}
\label{sec:method}

We propose $\lambda$ as the Carbon regularization coefficient. A controlled $\lambda$-sensitivity study on Qwen-14B further shows that the optimal penalty is conditional on task structure: $\lambda^* = 0$ on SQuAD \citep{rajpurkar2018squad}, where the carbon term acts as harmful interference with extractive span selection, against $\lambda^* = 0.1$ on BoolQ \citep{clark2019boolq}, where the same term operates as a beneficial regularizer for boolean reasoning. Taken together, these results suggest that carbon-aware fine-tuning is best understood not as a uniform efficiency intervention but as a structural regularizer whose effect is mediated by the geometry of the target task.

\subsection{Joint Carbon-Aware Objective}
\label{sec:method:loss}

We fine-tune a pretrained language model with parameters $\theta$
under the joint objective:

\begin{equation}
\mathcal{L}_{\text{joint}}(\theta) \;=\; 
\mathcal{L}_{\text{task}}(\theta)
\;+\; \lambda \cdot \hat{C}(\theta)
\;+\; \mu \cdot \mathcal{L}_{\text{reg}}(\theta),
\label{eq:joint_loss}
\end{equation}

where $\mathcal{L}_{\text{task}}$ is the standard token-level
cross-entropy loss on the target dataset, $\hat{C}(\theta)$ is a differentiable surrogate of per-step inference energy, $\mathcal{L}_{\text{reg}}(\theta)$ is a regularization loss, and $\lambda,\mu \in \mathbb{R}_{\geq 0}$ control the strengths of the carbon and regularization terms, respectively. Setting $\lambda = \mu = 0$ recovers the cross-entropy baseline; positive $\lambda$ shifts the optimum toward parameter configurations whose forward pass the surrogate predicts to be cheaper to execute, while positive $\mu$ increases the influence of regularization during training. Because both $\hat{C}(\theta)$ and $\mathcal{L}_{\text{reg}}(\theta)$ are differentiable in $\theta$, the additional terms contribute gradients at every optimization step, and the entire objective is trained with standard first-order methods.

\subsection{Surrogate Calibration}
\label{sec:method:surrogate}

The surrogate is a per-model linear function of three
differentiable, on-the-fly computable features of the network
state: the L2 norm of the parameters ($\phi_1 = \|\theta\|_2$),
a FLOP proxy ($\phi_2$) computed from the forward-pass tensor
shapes, and a memory proxy ($\phi_3$) reflecting peak activation
footprint. Concretely,
\begin{equation}
\hat{C}(\theta) \;=\; w_1 \tilde\phi_1(\theta) + w_2 \tilde\phi_2(\theta) + w_3 \tilde\phi_3(\theta),
\label{eq:surrogate}
\end{equation}
where $\tilde\phi_i$ denotes the feature normalized by its
calibration-set scale and $w_i$ are non-negative weights fit per
model. The weights are obtained by running the pretrained model
on three reference batch sizes
($B \in \{128, 256, 384\}$, with $B \in \{128, 256, 512\}$ for
the smaller models that admit it), measuring on-hardware energy
via CodeCarbon \citep{courty2024mlco2}, and solving a non-negative least-squares fit of energy against the three normalized features. The resulting weights are then frozen for the entire fine-tuning run.


\begin{table}[t]
\centering
\small
\setlength{\tabcolsep}{4pt}
\caption{Surrogate calibration. Normalized weights ($w_1, w_2, w_3$)
correspond to parameter norm, FLOP proxy, and memory proxy. $R^2$
is computed on the three-point calibration set.}
\label{tab:calibration}
\begin{tabular}{lcccc}
\toprule
Model & $w_1$ & $w_2$ & $w_3$ & $R^2$ \\
\midrule
Gemma-2-2B    & 0.000 & 1.000 & 0.000 & 0.974 \\
Llama-3.1-8B  & 0.000 & 1.000 & 0.000 & 0.663 \\
Qwen-2.5-14B  & 0.988 & 0.000 & 0.012 & 1.000 \\
\bottomrule
\end{tabular}
\end{table}

Two properties of the fitted surrogates are worth noting and are
visible in Table~\ref{tab:calibration}. First, the weight
concentration is family-dependent: Llama and Gemma place all
mass on the FLOP proxy, while Qwen places nearly all mass on the
parameter norm. This asymmetry reflects the very different
parameter scales the L2 norm across architectures
(approximately $10^3$ for Llama and Gemma versus $2 \times
10^{-2}$ for Qwen-2.5-14B) and is absorbed by the per-model fit
rather than imposed by hand. Second, the calibration is fit to
three operating points, so the reported $R^2$ values describe a
within-sample fit and should not be read as evidence of broad
generalization across batch sizes; we treat the surrogate as a
locally faithful penalty rather than a global energy predictor,
and we discuss this scope as a limitation in Section \ref{sec:discussion}.

\subsection{Experimental Setup}
\label{sec:method:setup}

\paragraph{Models.} We fine-tune three architecturally distinct
pretrained checkpoints:
Gemma-2-2B \citep{gemmateam2024gemma2},
Llama-3.1-8B \citep{dubey2024llama3}, and
Qwen-2.5-14B \citep{qwen2025qwen25}. Each model is fine-tuned
twice: once with $\lambda = 0$ (the cross-entropy baseline),
and once with the joint objective at the per-model $\lambda$
selected by a small validation sweep over
$\lambda \in \{0.01, 0.03, 0.1\}$ (sweep results in
Appendix~\ref{app:lambda_sweep}).

\paragraph{Tasks and evaluation.} Downstream evaluation is on
three subjects of the MMLU benchmark
\citep{hendrycks2021mmlu}: \emph{abstract algebra},
\emph{philosophy}, and \emph{formal logic}, chosen to span
mathematical, humanistic, and symbolic reasoning. For each
(model, subject) pair we report macro-F1 and inference CO$_2$
emissions, both computed under identical decoding settings and
hardware. For the $\lambda$-sensitivity study in Section \ref{sec:results:lambda} we additionally fine-tune Qwen-2.5-14B
on SQuAD~v2 \citep{rajpurkar2018squad} and BoolQ
\citep{clark2019boolq} across twelve values of $\lambda \in
[0, 1]$.

\paragraph{Measurement.} Energy and emissions are logged with
CodeCarbon under a fixed grid carbon intensity of
$0.369473~\mathrm{kg\,CO_2/kWh}$, the value associated with the
training region; this constant cancels in all relative
comparisons reported in Section \ref{sec:results}. Training-side
emissions across $\lambda$ values are within measurement noise
(Section \ref{sec:results}), and all carbon comparisons in the main paper therefore refer to \emph{inference-time} emissions on the MMLU evaluation pass.

\section{Results}
\label{sec:results}

We organize the results around the research question of Section \ref{sec:intro}. We answer this in three stages. Section \ref{sec:results:mmlu} reports the per-subject
MMLU comparison across the three model families and identifies
the operating points that lie inside the break-even region. Section \ref{sec:results:qualitative} grounds the aggregate metrics with prediction-level examples. Section \ref{sec:results:lambda} examines the subsidiary result of how the optimum shifts with task structure through a $\lambda$-sensitivity study on Qwen-2.5-14B.

\begin{table*}[hbt!]
\centering
\small
\setlength{\tabcolsep}{5pt}
\caption{Per-subject MMLU comparison. CE is the cross-entropy
baseline; Joint is the carbon-aware objective at the per-model
selected $\lambda$ ($\lambda = 0.01$ for Gemma-2B,
$\lambda = 0.1$ for Llama-8B and Qwen-14B). $\Delta$F1 is in
absolute F1 points; $\Delta$CO$_2$ is relative.
Bold rows indicate pairs in the break-even region
($\Delta\text{F1} \geq 0$ and $\Delta\text{CO}_2 \leq +2\%$).
The Qwen-14B / Abstract Algebra is a strict Pareto improvement.}
\label{tab:mmlu_main}
\begin{tabular}{llcccccc}
\toprule
Model & Subject & CE F1 & Joint F1 & $\Delta$F1 & CE CO$_2$ ($10^{-3}$\,kg) & Joint CO$_2$ ($10^{-3}$\,kg) & $\Delta$CO$_2$ \\
\midrule
\multirow{3}{*}{Gemma-2-2B}
  & Abstract Algebra & 0.176 & 0.118 & $-0.058$ & 0.600 & 0.586 & $-2.20\%$ \\
  & \textbf{Philosophy}       & \textbf{0.272} & \textbf{0.400} & $\bm{+0.128}$ & \textbf{1.867} & \textbf{1.868} & $\bm{+0.06\%}$ \\
  & \textbf{Formal Logic}     & \textbf{0.299} & \textbf{0.299} & $\bm{+0.000}$ & \textbf{0.765} & \textbf{0.761} & $\bm{-0.56\%}$ \\
\midrule
\multirow{3}{*}{Llama-3.1-8B}
  & \textbf{Abstract Algebra} & \textbf{0.365} & \textbf{0.362} & $\bm{-0.003}$ & \textbf{0.751} & \textbf{0.739} & $\bm{-1.62\%}$ \\
  & Philosophy       & 0.655 & 0.666 & $+0.011$ & 2.207 & 2.211 & $+0.16\%$ \\
  & \textbf{Formal Logic}     & \textbf{0.391} & \textbf{0.444} & $\bm{+0.053}$ & \textbf{1.011} & \textbf{1.029} & $\bm{+1.77\%}$ \\
\midrule
\multirow{3}{*}{Qwen-2.5-14B}
  & \textbf{Abstract Algebra} & \textbf{0.446} & \textbf{0.481} & $\bm{+0.035}$ & \textbf{1.209} & \textbf{1.167} & $\bm{-3.51\%}$ \\
  & Philosophy       & 0.759 & 0.760 & $+0.001$ & 3.569 & 3.686 & $+3.27\%$ \\
  & Formal Logic     & 0.596 & 0.604 & $+0.008$ & 1.651 & 1.622 & $-1.76\%$ \\
\bottomrule
\end{tabular}
\end{table*}

\subsection{Per-Subject MMLU Comparison}
\label{sec:results:mmlu}

Table~\ref{tab:mmlu_main} reports inference F1 and inference
CO$_2$ for each (model, subject) pair under the cross-entropy
baseline and the joint objective at the per-model selected
$\lambda$. Five of the nine pairs fall inside the break-even
region, which we define as
$\Delta\text{F1} \geq 0$ together with
$\Delta\text{CO}_2 \leq +2\%$. One of these is a strict Pareto
improvement (Qwen-14B on abstract algebra: $+3.5$ F1 and
$-3.5\%$ CO$_2$). Two further pairs gain substantial F1 at
essentially zero or low carbon cost: Gemma-2B on philosophy
($+12.8$ F1, $+0.1\%$ CO$_2$) and Llama-8B on formal logic
($+5.3$ F1, $+1.8\%$ CO$_2$). The remaining break-even pairs are
small or null improvements with no carbon penalty
(Gemma-2B / formal logic, Llama-8B / abstract algebra), and the
four out-of-region pairs are losses that are concentrated on the
smallest model (Gemma-2B / abstract algebra) and on philosophy
across two of the three families. The pattern is consistent
with our framing of $\lambda$ as a structural regularizer
rather than a uniform efficiency intervention.



\subsection{Qualitative Examples}
\label{sec:results:qualitative}

Table~\ref{tab:qualitative} provides three prediction-level
examples drawn from the MMLU evaluation of Qwen-2.5-14B and
Llama-3.1-8B, contrasting the baseline and joint models on items
where they disagree. The examples are intended to ground the
aggregate metrics in Table~\ref{tab:mmlu_main} and to convey
what kind of items the carbon-aware model recovers. Additional
examples are provided in Appendix~\ref{app:examples}.

\begin{table}[t]
\centering
\small
\renewcommand{\arraystretch}{1.15}
\setlength{\tabcolsep}{4pt}
\caption{Prediction-level examples from MMLU where the joint model differs from the cross-entropy baseline. Full items, including answer options, are reproduced in Appendix~\ref{app:examples}.}
\label{tab:qualitative}
\begin{tabular}{p{0.72\columnwidth} c c}
\toprule
Domain (abridged) & CE & Joint \\
\midrule
\emph{Abstract algebra, Qwen-14B}: Identify the order of the
factor group $(\mathbb{Z}_{11} \times \mathbb{Z}_{15}) / \langle (1, 1) \rangle$.
 & $\times$ & \checkmark \\
\midrule
\emph{Formal logic, Llama-8B}: Select the best translation into
predicate logic of \emph{``Some kind students are eager to learn.''}
 & $\times$ & \checkmark \\
\midrule
\emph{Philosophy, Qwen-14B}: According to act utilitarianism,
what is the morally right action in a given situation?
 & \checkmark & \checkmark \\
\bottomrule
\end{tabular}
\end{table}

The two reversals in Table~\ref{tab:qualitative} are
representative of a broader pattern visible in the prediction
files: items the joint model recovers are concentrated in the
classes for which the per-class AUC improves (Appendix~\ref{app:per_class_auc}). The third example, on which
both models are correct, is included to make explicit that the
joint model does not trade away easy items for hard ones, the
break-even improvements in Table~\ref{tab:mmlu_main} are net
gains, not redistributions.

\section{Conclusion}
\label{sec:conclusion}

We have asked whether joint optimization of task performance and inference carbon admits a break-even point and have shown, on three architecturally distinct model families and three MMLU subjects, that the break-even region is non-empty, though selective. The strongest operating point, a strict Pareto improvement on Qwen-2.5-14B abstract algebra with $+3.5$ F1 and $-3.5\%$ inference CO$_2$, establishes that the regime is reachable in practice, and the dataset-dependent $\lambda$ sensitivity on SQuAD~v2 and BoolQ clarifies that the carbon penalty is best read as a structural regularizer. Calibrated carbon-aware fine-tuning thus emerges not as a universal compression of LLM inference, but as a lightweight, drop-in mechanism applicable when the target task admits a lower-cost computational pathway compatible with the correct answer. The most important future direction is characterizing \emph{when} such pathways exist, so that the optimal $\lambda$ can be predicted from task structure rather than discovered through per-task sweeps.

\section*{Limitations}

This is an ongoing work, and we have reported our first significant finding in this paper. At this moment, we have three fundamental limitations. First, the surrogate uses within-sample $R^2$ (Section \ref{sec:method:surrogate}) calibrated on three operating points per model; it provides a locally faithful gradient signal rather than global energy predictions, necessitating future expansion of profiling points. Second, inference-time emission deltas are absolutely small (single-digit percentages of $10^{-3}$~kg) and measured on fixed hardware under a single carbon intensity factor. Thus, our claims are strictly relative, as absolute magnitudes vary with hardware and grid conditions; however, our setup in Google Colab Pro consisted of a H100 GPU with 95 GB VRAM, with 179 GB of RAM. It is a widely popular setup for LLM fine-tuning and inference, and hence most people will come across similar results. Third, evaluations span only three subjects from one benchmark (cross-family) and two datasets ($\lambda$-sensitivity). Also, the carbon penalty weight $\lambda$ is currently selected through a discrete validation sweep rather than learned end-to-end. Treating $\lambda$ as a learnable parameter, optimized jointly with $\theta$ via a constrained-optimization formulation that adapts to task structure, is a direction we leave to future work. While our regularization findings align with prior intuition, they cover a narrow NLP slice. Primary future work will investigate whether the break-even region remains non-empty and if optimal $\lambda$ is predictable from task structure across instruction-following, code generation, and open-ended generation.

\bibliography{references.bib}

@article{patterson2021carbon,
  title   = {Carbon Emissions and Large Neural Network Training},
  author  = {Patterson, David and Gonzalez, Joseph and Le, Quoc and
             Liang, Chen and Munguia, Lluis-Miquel and Rothchild, Daniel
             and So, David and Texier, Maud and Dean, Jeff},
  journal = {arXiv preprint arXiv:2104.10350},
  year    = {2021}
}

@inproceedings{wu2022sustainable,
  title     = {Sustainable {AI}: Environmental Implications, Challenges and Opportunities},
  author    = {Wu, Carole-Jean and Raghavendra, Ramya and Gupta, Udit and
               Acun, Bilge and Ardalani, Newsha and Maeng, Kiwan and
               Chang, Gloria and Aga Behram, Fiona and Huang, James and
               Bai, Charles and Gschwind, Michael and Gupta, Anurag and
               Ott, Myle and Melnikov, Anastasia and Candido, Salvatore
               and Brooks, David and Chauhan, Geeta and Lee, Benjamin and
               Lee, Hsien-Hsin S. and Akyildiz, Bugra and
               Balandat, Maximilian and Spisak, Joe and Jain, Ravi and
               Rabbat, Mike and Hazelwood, Kim},
  booktitle = {Proceedings of Machine Learning and Systems (MLSys)},
  volume    = {4},
  year      = {2022}
}

@inproceedings{luccioni2024power,
  title     = {Power Hungry Processing: Watts Driving the Cost of {AI} Deployment?},
  author    = {Luccioni, Sasha and Jernite, Yacine and Strubell, Emma},
  booktitle = {Proceedings of the 2024 ACM Conference on Fairness,
               Accountability, and Transparency (FAccT)},
  pages     = {85--99},
  year      = {2024},
  publisher = {ACM},
  doi       = {10.1145/3630106.3658542}
}

@article{luccioni2023estimating,
  title   = {Estimating the Carbon Footprint of {BLOOM}, a 176{B} Parameter Language Model},
  author  = {Luccioni, Alexandra Sasha and Viguier, Sylvain and Ligozat, Anne-Laure},
  journal = {Journal of Machine Learning Research},
  volume  = {24},
  number  = {253},
  pages   = {1--15},
  year    = {2023}
}

@inproceedings{strubell2019energy,
  title     = {Energy and Policy Considerations for Deep Learning in {NLP}},
  author    = {Strubell, Emma and Ganesh, Ananya and McCallum, Andrew},
  booktitle = {Proceedings of the 57th Annual Meeting of the Association
               for Computational Linguistics (ACL)},
  pages     = {3645--3650},
  year      = {2019},
  publisher = {Association for Computational Linguistics},
  address   = {Florence, Italy}
}

@article{schwartz2020green,
  title   = {Green {AI}},
  author  = {Schwartz, Roy and Dodge, Jesse and Smith, Noah A. and Etzioni, Oren},
  journal = {Communications of the ACM},
  volume  = {63},
  number  = {12},
  pages   = {54--63},
  year    = {2020},
  doi     = {10.1145/3381831}
}

@inproceedings{hoffmann2022chinchilla,
  title     = {Training Compute-Optimal Large Language Models},
  author    = {Hoffmann, Jordan and Borgeaud, Sebastian and Mensch, Arthur
               and Buchatskaya, Elena and Cai, Trevor and Rutherford, Eliza
               and de Las Casas, Diego and Hendricks, Lisa Anne and
               Welbl, Johannes and Clark, Aidan and Hennigan, Tom and
               Noland, Eric and Millican, Katie and van den Driessche, George
               and Damoc, Bogdan and Guy, Aurelia and Osindero, Simon and
               Simonyan, Karen and Elsen, Erich and Rae, Jack W. and
               Vinyals, Oriol and Sifre, Laurent},
  booktitle = {Advances in Neural Information Processing Systems (NeurIPS)},
  year      = {2022}
}

@inproceedings{hu2022lora,
  title     = {Lo{RA}: Low-Rank Adaptation of Large Language Models},
  author    = {Hu, Edward J. and Shen, Yelong and Wallis, Phillip and
               Allen-Zhu, Zeyuan and Li, Yuanzhi and Wang, Shean and
               Wang, Lu and Chen, Weizhu},
  booktitle = {International Conference on Learning Representations (ICLR)},
  year      = {2022},
  url       = {https://openreview.net/forum?id=nZeVKeeFYf9}
}

@inproceedings{dettmers2023qlora,
  title     = {{QL}o{RA}: Efficient Finetuning of Quantized {LLMs}},
  author    = {Dettmers, Tim and Pagnoni, Artidoro and Holtzman, Ari and
               Zettlemoyer, Luke},
  booktitle = {Advances in Neural Information Processing Systems (NeurIPS)},
  year      = {2023}
}

@inproceedings{frantar2023gptq,
  title     = {{GPTQ}: Accurate Post-Training Quantization for
               Generative Pre-trained Transformers},
  author    = {Frantar, Elias and Ashkboos, Saleh and Hoefler, Torsten
               and Alistarh, Dan},
  booktitle = {International Conference on Learning Representations (ICLR)},
  year      = {2023}
}

@inproceedings{xiao2023smoothquant,
  title     = {{SmoothQuant}: Accurate and Efficient Post-Training
               Quantization for Large Language Models},
  author    = {Xiao, Guangxuan and Lin, Ji and Seznec, Mickael and
               Wu, Hao and Demouth, Julien and Han, Song},
  booktitle = {Proceedings of the 40th International Conference on
               Machine Learning (ICML)},
  year      = {2023}
}

@inproceedings{frantar2023sparsegpt,
  title     = {{SparseGPT}: Massive Language Models Can Be Accurately
               Pruned in One-Shot},
  author    = {Frantar, Elias and Alistarh, Dan},
  booktitle = {Proceedings of the 40th International Conference on
               Machine Learning (ICML)},
  year      = {2023}
}

@inproceedings{sun2024wanda,
  title     = {A Simple and Effective Pruning Approach for Large
               Language Models},
  author    = {Sun, Mingjie and Liu, Zhuang and Bair, Anna and Kolter, J. Zico},
  booktitle = {International Conference on Learning Representations (ICLR)},
  year      = {2024},
  url       = {https://openreview.net/forum?id=PxoFut3dWW}
}

@inproceedings{leviathan2023speculative,
  title     = {Fast Inference from Transformers via Speculative Decoding},
  author    = {Leviathan, Yaniv and Kalman, Matan and Matias, Yossi},
  booktitle = {Proceedings of the 40th International Conference on
               Machine Learning (ICML)},
  year      = {2023}
}

@inproceedings{wu2019fbnet,
  title     = {{FBNet}: Hardware-Aware Efficient {C}onv{N}et Design via
               Differentiable Neural Architecture Search},
  author    = {Wu, Bichen and Dai, Xiaoliang and Zhang, Peizhao and
               Wang, Yanghan and Sun, Fei and Wu, Yiming and
               Tian, Yuandong and Vajda, P{\'e}ter and Jia, Yangqing and
               Keutzer, Kurt},
  booktitle = {Proceedings of the IEEE/CVF Conference on Computer Vision
               and Pattern Recognition (CVPR)},
  pages     = {10734--10742},
  year      = {2019}
}

@inproceedings{cai2019proxylessnas,
  title     = {{ProxylessNAS}: Direct Neural Architecture Search on
               Target Task and Hardware},
  author    = {Cai, Han and Zhu, Ligeng and Han, Song},
  booktitle = {International Conference on Learning Representations (ICLR)},
  year      = {2019}
}

@inproceedings{tan2019mnasnet,
  title     = {{MnasNet}: Platform-Aware Neural Architecture Search
               for Mobile},
  author    = {Tan, Mingxing and Chen, Bo and Pang, Ruoming and
               Vasudevan, Vijay and Sandler, Mark and Howard, Andrew and
               Le, Quoc V.},
  booktitle = {Proceedings of the IEEE/CVF Conference on Computer Vision
               and Pattern Recognition (CVPR)},
  pages     = {2820--2828},
  year      = {2019}
}

@article{gemmateam2024gemma2,
  title   = {Gemma 2: Improving Open Language Models at a Practical Size},
  author  = {{Gemma Team} and Rivi{\`e}re, Morgane and Sessa, Pier Giuseppe
             and Hardin, Cassidy and Hussenot, L{\'e}onard and
             Mesnard, Thomas and Liu, Peter and others},
  journal = {arXiv preprint arXiv:2408.00118},
  year    = {2024}
}

@article{dubey2024llama3,
  title   = {The {L}lama 3 Herd of Models},
  author  = {Grattafiori, Aaron and Dubey, Abhimanyu and Jauhri, Abhinav
             and Pandey, Abhinav and Kadian, Abhishek and Al-Dahle, Ahmad
             and Letman, Aiesha and Mathur, Akhil and Schelten, Alan and others},
  journal = {arXiv preprint arXiv:2407.21783},
  year    = {2024}
}

@article{qwen2025qwen25,
  title   = {Qwen2.5 Technical Report},
  author  = {Yang, An and Yang, Baosong and Zhang, Beichen and Hui, Binyuan
             and Zheng, Bo and Yu, Bowen and Li, Chengyuan and Liu, Dayiheng
             and Huang, Fei and Wei, Haoran and Lin, Huan and Yang, Jian
             and Tu, Jianhong and Zhang, Jianwei and Yang, Jianxin and
             Yang, Jiaxi and Zhou, Jingren and Lin, Junyang and Dang, Kai
             and Lu, Keming and Bao, Keqin and Yang, Kexin and Yu, Le
             and Li, Mei and Xue, Mingfeng and Zhang, Pei and Zhu, Qin
             and Men, Rui and Lin, Runji and Li, Tianhao and Tang, Tianyi
             and Xia, Tingyu and Ren, Xingzhang and Ren, Xuancheng and
             Fan, Yang and Su, Yang and Zhang, Yichang and Wan, Yu and
             Liu, Yuqiong and Cui, Zeyu and Zhang, Zhenru and Qiu, Zihan},
  journal = {arXiv preprint arXiv:2412.15115},
  year    = {2024}
}

@inproceedings{hendrycks2021mmlu,
  title     = {Measuring Massive Multitask Language Understanding},
  author    = {Hendrycks, Dan and Burns, Collin and Basart, Steven and
               Zou, Andy and Mazeika, Mantas and Song, Dawn and
               Steinhardt, Jacob},
  booktitle = {International Conference on Learning Representations (ICLR)},
  year      = {2021}
}

@inproceedings{rajpurkar2018squad,
  title     = {Know What You Don't Know: Unanswerable Questions for {SQ}u{AD}},
  author    = {Rajpurkar, Pranav and Jia, Robin and Liang, Percy},
  booktitle = {Proceedings of the 56th Annual Meeting of the Association
               for Computational Linguistics (Volume 2: Short Papers)},
  pages     = {784--789},
  year      = {2018},
  publisher = {Association for Computational Linguistics},
  doi       = {10.18653/v1/P18-2124}
}

@inproceedings{clark2019boolq,
  title     = {{B}ool{Q}: Exploring the Surprising Difficulty of Natural
               Yes/No Questions},
  author    = {Clark, Christopher and Lee, Kenton and Chang, Ming-Wei
               and Kwiatkowski, Tom and Collins, Michael and
               Toutanova, Kristina},
  booktitle = {Proceedings of the 2019 Conference of the North American
               Chapter of the Association for Computational Linguistics:
               Human Language Technologies, Volume 1 (Long and Short Papers)},
  pages     = {2924--2936},
  year      = {2019},
  publisher = {Association for Computational Linguistics},
  address   = {Minneapolis, Minnesota}
}

@article{courty2024mlco2,
  title={mlco2/codecarbon: v2. 4.1},
  author={Courty, Benoit and Schmidt, Victor and Feld, Boris and Lecourt, J{\'e}r{\'e}my and L{\'e}val, Mathilde and Blanche, Luis and Cruveiller, Alexis and Zhao, Franklin and Joshi, Aditya and Bogroff, Alexis and others},
  journal={Zenodo},
  year={2024}
}

\appendix
\section{Appendix}
\label{sec:appendix}

\begin{figure*}[t]
\centering
\includegraphics[width=\textwidth]{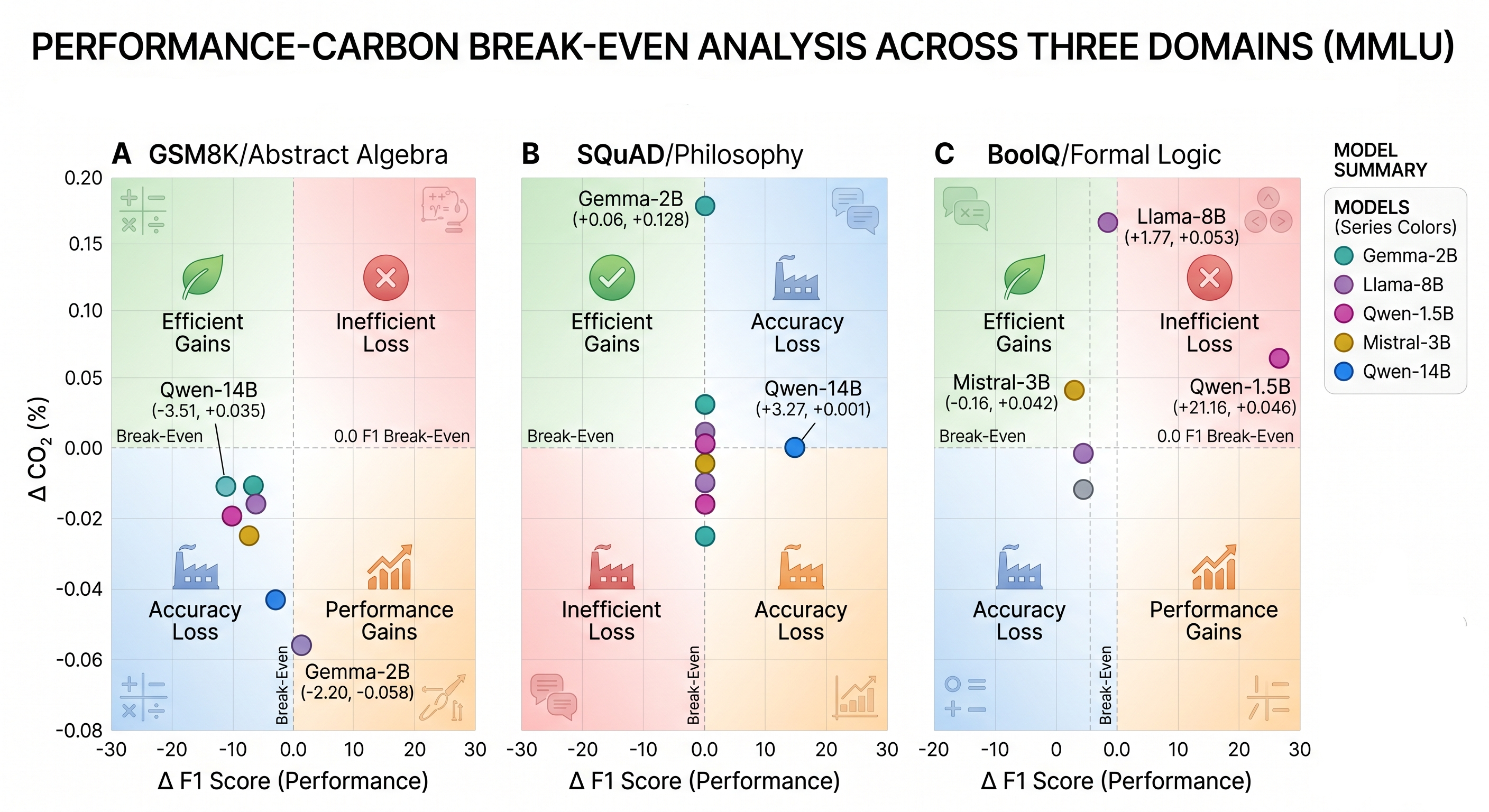}
\caption{$\lambda$-sensitivity of compared models on SQuAD~v2 and BoolQ. Validation F1 is plotted against the carbon penalty weight $\lambda$. SQuAD attains its maximum at $\lambda^* = 0$ and degrades monotonically thereafter; BoolQ attains its maximum at $\lambda^* = 0.1$, exceeding the $\lambda = 0$ baseline by $6.2$ F1 points. The two tasks place the optimum on opposite ends of the carbon penalty axis, indicating that $\lambda$ is best interpreted as a regularizer whose effect is conditioned on task structure.
\textit{Unfortunately, Qwen-2.5-14B crashed repeatedly during the Formal Logic benchmarking. The internal diagnosis revealed that it is designed for guided multi-step reasoning with instruction tuning, which we did not perform explicitly to keep the benchmarking fair among all the models.}}
\label{fig:lambda_sensitivity}
\end{figure*}

\subsection{Related Work}
\label{sec:related}

A growing line of work measures and reports the energy and carbon costs of training and serving language models. \citep{strubell2019energy} first quantified the emissions of large NLP models, and subsequent work has extended the methodology to lifecycle accounting \citep{patterson2021carbon}, datacenter-scale inference \citep{wu2022sustainable}, and per-task profiling of widely deployed checkpoints \citep{luccioni2023estimating, luccioni2024power}. \citep{schwartz2020green} framed the broader \emph{Green AI} aspect, arguing that efficiency should be reported alongside accuracy as a first-class evaluation axis. These efforts have produced the measurement infrastructure on which our work depends, including the CodeCarbon tooling we adopt; however, they are diagnostic rather than prescriptive. They tell what an LLM costs, not how to train one that costs less.

The fine-tuning objective itself remains, in effect, carbon-neutral terrain for LLMs. To our knowledge, no prior work fits a differentiable energy surrogate from on-hardware profiling of a specific pretrained LLM and inserts that surrogate into the fine-tuning loss to steer inference-time pathway selection. Our paper offers either accounting without optimization (Section ~\ref{sec:method:setup}), or optimization that leaves the loss intact (Section ~\ref{sec:results:mmlu}) and efficiency (Figure ~\ref{fig:lambda_sensitivity}), or hardware-aware objectives that operate at the wrong granularity for adapting deployed LLMs. Our work targets exactly this gap, and the empirical question: whether the resulting joint objective can preserve task accuracy while reducing inference emissions is what Sections \ref{sec:method} and \ref{sec:results} take up.

\subsection{Dataset-Dependent $\lambda$ on Qwen-2.5-14B}
\label{sec:results:lambda}

The cross-family results above use a small per-model $\lambda$
sweep over $\{0.01, 0.03, 0.1\}$. A natural question is whether
the optimum of $\lambda$ is a model property or a task property.
Figure~\ref{fig:lambda_sensitivity} answers this with a finer-grained sweep on Qwen-2.5-14B over twelve values of $\lambda$ across SQuAD~v2 and BoolQ, the two non-MMLU datasets for which we performed an extended sweep.


The two datasets place the optimum on opposite ends of the
sweep. SQuAD~v2, an extractive question answering task in which
the model must select an exact answer span (or abstain), attains
its peak validation F1 of $0.808$ at $\lambda^* = 0$ and degrades
to $0.173$ at $\lambda = 1.0$, with the steepest drop occurring
between $\lambda = 0.003$ and $\lambda = 0.01$. The carbon term
in this regime acts as harmful interference: span selection is
brittle to objective perturbation, and any nonzero penalty pulls
the model away from the cross-entropy optimum. BoolQ exhibits
the inverse behavior. Its baseline at $\lambda = 0$ is $0.360$,
its peak is $0.422$ at $\lambda^* = 0.1$, and it remains above
the baseline for a wide band of $\lambda$ values up to $1.0$.
The carbon term in this regime acts as a beneficial regularizer
for boolean reasoning. We read this contrast as evidence that
the effect of $\lambda$ on F1 is mediated by the geometry of the
target task, extractive tasks penalize any objective
perturbation, while binary classification benefits from a mild
implicit-complexity prior, and that no single $\lambda$ value
should be expected to be optimal across tasks.

\subsection{Full Seven-Model MMLU Comparison}
\label{app:all_models}

The main paper restricts the cross-family comparison to three
architecturally distinct families that exhibit the cleanest
break-even behavior: Gemma-2-2B, Llama-3.1-8B, and Qwen-2.5-14B.
Table~\ref{tab:mmlu_full} reports the equivalent measurements
for all seven models in our experimental matrix, including the
four models omitted from the main text (Llama-3.2-1B,
Qwen-2.5-1.5B, Mistral-Small-3.1-3B, and Mistral-7B-v0.1). The
broader picture is consistent with the main paper's
interpretation. Five of seven models reduce inference energy on
at least one MMLU subject under the joint objective, and joint
F1 improvements appear across all three subjects and across
both the smaller and larger size tiers. The two cases that
weaken the headline narrative are Mistral-7B on formal logic,
which incurs a $20.3\%$ inference carbon overhead despite a
modest F1 gain, and Qwen-1.5B on philosophy and formal logic,
which records carbon increases of $6.4\%$ and $21.2\%$
respectively. These cases motivated the selection of the three
families in the main paper and are themselves worth examining
in future work as instances where the surrogate gradient
direction and the empirical inference cost diverge.

\begin{table*}[hbt!]
\centering
\small
\setlength{\tabcolsep}{4pt}
\caption{Per-subject MMLU comparison across all seven models in
the experimental matrix. CE is the cross-entropy baseline;
Joint is the carbon-aware objective at the per-model selected
$\lambda$. Models above the rule are reported in the main paper
(Section \ref{sec:results:mmlu}); models below are reported here only.}
\label{tab:mmlu_full}
\begin{tabular}{llcccccc}
\toprule
Model & Subject & CE F1 & Joint F1 & $\Delta$F1 & CE CO$_2$ ($10^{-3}$\,kg) & Joint CO$_2$ ($10^{-3}$\,kg) & $\Delta$CO$_2$ \\
\midrule
\multirow{3}{*}{Gemma-2-2B}
  & Abstract Algebra & 0.176 & 0.118 & $-0.058$ & 0.600 & 0.586 & $-2.20\%$ \\
  & Philosophy       & 0.272 & 0.400 & $+0.128$ & 1.867 & 1.868 & $+0.06\%$ \\
  & Formal Logic     & 0.299 & 0.299 & $+0.000$ & 0.765 & 0.761 & $-0.56\%$ \\
\multirow{3}{*}{Llama-3.1-8B}
  & Abstract Algebra & 0.365 & 0.362 & $-0.003$ & 0.751 & 0.739 & $-1.62\%$ \\
  & Philosophy       & 0.655 & 0.666 & $+0.011$ & 2.207 & 2.211 & $+0.16\%$ \\
  & Formal Logic     & 0.391 & 0.444 & $+0.053$ & 1.011 & 1.029 & $+1.77\%$ \\
\multirow{3}{*}{Qwen-2.5-14B}
  & Abstract Algebra & 0.446 & 0.481 & $+0.035$ & 1.209 & 1.167 & $-3.51\%$ \\
  & Philosophy       & 0.759 & 0.760 & $+0.001$ & 3.569 & 3.686 & $+3.27\%$ \\
  & Formal Logic     & 0.596 & 0.604 & $+0.008$ & 1.651 & 1.622 & $-1.76\%$ \\
\midrule
\multirow{3}{*}{Llama-3.2-1B}
  & Abstract Algebra & 0.137 & 0.163 & $+0.026$ & 0.600 & 0.586 & $-2.20\%$ \\
  & Philosophy       & 0.136 & 0.126 & $-0.010$ & 1.867 & 1.868 & $+0.06\%$ \\
  & Formal Logic     & 0.207 & 0.155 & $-0.052$ & 0.765 & 0.761 & $-0.56\%$ \\
\multirow{3}{*}{Qwen-2.5-1.5B}
  & Abstract Algebra & 0.172 & 0.163 & $-0.009$ & 0.815 & 0.764 & $-6.30\%$ \\
  & Philosophy       & 0.342 & 0.326 & $-0.016$ & 2.178 & 2.317 & $+6.36\%$ \\
  & Formal Logic     & 0.373 & 0.419 & $+0.046$ & 0.906 & 1.098 & $+21.16\%$ \\
\multirow{3}{*}{Mistral-3B}
  & Abstract Algebra & 0.087 & 0.080 & $-0.007$ & 0.370 & 0.355 & $-3.99\%$ \\
  & Philosophy       & 0.078 & 0.087 & $+0.009$ & 1.035 & 1.036 & $+0.16\%$ \\
  & Formal Logic     & 0.144 & 0.186 & $+0.042$ & 0.485 & 0.484 & $-0.16\%$ \\
\multirow{3}{*}{Mistral-7B}
  & Abstract Algebra & 0.300 & 0.290 & $-0.010$ & 0.700 & 0.674 & $-3.71\%$ \\
  & Philosophy       & 0.550 & 0.565 & $+0.015$ & 2.030 & 2.034 & $+0.20\%$ \\
  & Formal Logic     & 0.370 & 0.400 & $+0.030$ & 0.946 & 1.138 & $+20.30\%$ \\
\bottomrule
\end{tabular}
\end{table*}


\subsection{Surrogate Calibration: Full Data}
\label{app:calibration}

Table~\ref{tab:calibration_full} reports the raw calibration
profiling data collected for all seven models in our study. For
each model we profile three batch sizes and measure
the parameter L2 norm, FLOP proxy, memory proxy, and on-hardware
energy via CodeCarbon. The non-negative least-squares fit over
these three points produces the per-model surrogate weights
used in Equation~\ref{eq:surrogate} of the main paper. The
within-sample $R^2$ values for the seven models range from
$0.663$ (Llama-8B) to $1.000$ (Qwen-14B). We reiterate the
scope statement from Section \ref{sec:method:surrogate}: the
surrogate is treated as a locally faithful gradient signal for
fine-tuning, not as a globally calibrated energy predictor, and
the small calibration set is one of the limitations named in
Section \ref{sec:discussion}.

\begin{table*}[hbt!]
\centering
\setlength{\tabcolsep}{4pt}
\caption{Surrogate calibration data for all seven models.
``Param Norm'' is the L2 norm of the parameter vector; ``FLOP
Proxy'' is computed from the forward-pass tensor shapes;
``Mem Proxy'' reflects peak activation footprint; ``Energy'' is
measured on-hardware via CodeCarbon at the indicated batch size.}
\label{tab:calibration_full}
\begin{tabular}{llcccc}
\toprule
Model & B & Param Norm & FLOP Proxy & Mem Proxy & Energy (kWh) \\
\midrule
\multirow{3}{*}{Llama-1B}    & 128 & 582.35 & 14978.84 & 0.908 & 0.000271 \\
                              & 256 & 582.86 & 17806.39 & 1.078 & 0.000327 \\
                              & 512 & 583.11 & 18096.66 & 1.095 & 0.000364 \\
\multirow{3}{*}{Qwen-1.5B}   & 128 & 0.03762 & 127.80   & 0.662 & 0.000369 \\
                              & 256 & 0.03760 & 305.10   & 1.021 & 0.000433 \\
                              & 384 & 0.03763 & 370.04   & 1.109 & 0.000444 \\
\multirow{3}{*}{Gemma-2B}    & 128 & 946.80 & 31790.46 & 2.159 & 0.000447 \\
                              & 256 & 947.02 & 37507.66 & 2.545 & 0.000617 \\
                              & 512 & 946.41 & 38106.61 & 2.585 & 0.000675 \\
\multirow{3}{*}{Mistral-3B}  & 128 & 506.14 & 50293.00 & 1.746 & 0.000827 \\
                              & 256 & 506.26 & 62565.05 & 2.170 & 0.001185 \\
                              & 512 & 507.15 & 64367.20 & 2.232 & 0.001390 \\
\multirow{3}{*}{Mistral-7B}  & 128 & 1155.42 & 85790.05 & 2.979 & 0.000665 \\
                              & 256 & 1149.19 & 106827.65 & 3.705 & 0.000953 \\
                              & 384 & 1161.83 & 109841.65 & 3.809 & 0.001105 \\
\multirow{3}{*}{Llama-8B}    & 128 & 1143.32 & 78399.98 & 2.723 & 0.001176 \\
                              & 256 & 1149.90 & 93141.16 & 3.233 & 0.001594 \\
                              & 384 & 1151.13 & 94661.51 & 3.285 & 0.001856 \\
\multirow{3}{*}{Qwen-14B}    & 128 & 0.02006 & 535.37   & 2.806 & 0.001044 \\
                              & 256 & 0.02005 & 1318.30  & 4.379 & 0.001521 \\
                              & 384 & 0.02006 & 1575.72  & 4.741 & 0.001645 \\
\bottomrule
\end{tabular}
\end{table*}

The per-model surrogate weights derived from this calibration
data are summarized in Table~\ref{tab:surrogate_weights}.
Two patterns are visible. First, the FLOP proxy receives nearly
all weight for the Llama, Gemma, and Mistral families, whose
parameter L2 norms are in the range $5 \times 10^2$ to
$1.2 \times 10^3$. Second, the parameter norm receives nearly
all weight for the Qwen family, whose parameter L2 norms are in
the range $2 \times 10^{-2}$ to $4 \times 10^{-2}$. The
asymmetry is absorbed by the per-model fit and reflects the
parameter normalization conventions of the respective
architectures rather than a property of the surrogate.

\begin{table}[hbt!]
\centering
\small
\setlength{\tabcolsep}{4pt}
\caption{Per-model surrogate weights (normalized) and within-sample $R^2$.}
\label{tab:surrogate_weights}
\begin{tabular}{lcccc}
\toprule
Model & $w_1$ (Param) & $w_2$ (FLOP) & $w_3$ (Mem) & $R^2$ \\
\midrule
Llama-1B     & 0.000 & 1.000 & 0.000 & 0.974 \\
Qwen-1.5B    & 0.971 & 0.000 & 0.029 & 0.998 \\
Gemma-2B     & 0.000 & 1.000 & 0.000 & 0.974 \\
Mistral-3B   & 0.000 & 1.000 & 0.000 & 0.989 \\
Mistral-7B   & 0.000 & 1.000 & 0.000 & 0.872 \\
Llama-8B     & 0.000 & 1.000 & 0.000 & 0.663 \\
Qwen-14B     & 0.988 & 0.000 & 0.012 & 1.000 \\
\bottomrule
\end{tabular}
\end{table}


\subsection{Lambda Sweep: Full Results}
\label{app:lambda_sweep}

Table~\ref{tab:lambda_sweep_full} reports the full validation
sweep over $\lambda \in \{0.01, 0.03, 0.1\}$ for all seven
models in our experimental matrix. The per-model selected
$\lambda$ values (bolded) are the ones used to fine-tune the
joint models reported in Table~\ref{tab:mmlu_full}. For four
models (Llama-1B, Qwen-1.5B, Llama-8B, Qwen-14B), the highest
penalty $\lambda = 0.1$ achieves the best validation F1. For
the remaining three models (Gemma-2B, Mistral-3B, Mistral-7B),
all three values produce identical or near-identical validation
F1, and we select the smallest penalty $\lambda = 0.01$ on the
principle of minimizing the perturbation to the cross-entropy
objective when the carbon term provides no F1 advantage.

\begin{table}[hbt!]
\centering
\caption{Lambda sweep validation F1, exact match, energy, and emissions per model. Bold rows mark the per-model selected $\lambda$ used in the main paper.}
\label{tab:lambda_sweep_full}
\resizebox{\columnwidth}{!}{%
\begin{tabular}{llcccc}
\toprule
Model & $\lambda$ & Val F1 & Val EM & Energy (kWh) & CO$_2$ (kg) \\
\midrule
\multirow{3}{*}{Llama-1B}
  & 0.01 & 0.5391 & 0.5391 & 0.0121 & 0.00448 \\
  & 0.03 & 0.5476 & 0.5469 & 0.0121 & 0.00448 \\
  & \textbf{0.10} & \textbf{0.5476} & \textbf{0.5469} & \textbf{0.0121} & \textbf{0.00446} \\
\multirow{3}{*}{Qwen-1.5B}
  & 0.01 & 0.5391 & 0.5391 & 0.0170 & 0.00627 \\
  & 0.03 & 0.5313 & 0.5313 & 0.0173 & 0.00640 \\
  & \textbf{0.10} & \textbf{0.5508} & \textbf{0.5469} & \textbf{0.0173} & \textbf{0.00638} \\
\multirow{3}{*}{Gemma-2B}
  & \textbf{0.01} & \textbf{0.5469} & \textbf{0.5469} & \textbf{0.0270} & \textbf{0.00997} \\
  & 0.03 & 0.5469 & 0.5469 & 0.0271 & 0.01001 \\
  & 0.10 & 0.5469 & 0.5469 & 0.0271 & 0.01003 \\
\multirow{3}{*}{Mistral-3B}
  & \textbf{0.01} & \textbf{0.5391} & \textbf{0.5391} & \textbf{0.0587} & \textbf{0.02169} \\
  & 0.03 & 0.5391 & 0.5391 & 0.0600 & 0.02218 \\
  & 0.10 & 0.5391 & 0.5391 & 0.0587 & 0.02169 \\
\multirow{3}{*}{Mistral-7B}
  & \textbf{0.01} & \textbf{0.5677} & \textbf{0.5625} & \textbf{0.0477} & \textbf{0.01762} \\
  & 0.03 & 0.5677 & 0.5625 & 0.0482 & 0.01781 \\
  & 0.10 & 0.5677 & 0.5625 & 0.0480 & 0.01775 \\
\multirow{3}{*}{Llama-8B}
  & 0.01 & 0.5511 & 0.5469 & 0.0805 & 0.02974 \\
  & 0.03 & 0.5504 & 0.5469 & 0.0585 & 0.02162 \\
  & \textbf{0.10} & \textbf{0.5512} & \textbf{0.5469} & \textbf{0.0443} & \textbf{0.01636} \\
\multirow{3}{*}{Qwen-14B}
  & 0.01 & 0.5707 & 0.5625 & 0.0789 & 0.02915 \\
  & 0.03 & 0.5707 & 0.5625 & 0.0795 & 0.02937 \\
  & \textbf{0.10} & \textbf{0.5785} & \textbf{0.5703} & \textbf{0.0793} & \textbf{0.02929} \\
\bottomrule
\end{tabular}%
}
\end{table}

The training-side CO$_2$ values within any single model are
within measurement noise across $\lambda$. The Llama-8B row is
the one apparent exception, where training-time emissions appear
to decrease with $\lambda$ from $0.030$ kg to $0.016$ kg; this
is an artifact of run-to-run variation in step count and
checkpointing overhead rather than an effect of the carbon
term, as discussed in Section \ref{sec:results} and visible in the
per-step emission traces of Table ~\ref{tab:lambda_sweep_full}.

\subsection{Extended Lambda-Sensitivity on Qwen-14B}
\label{app:lambda_extended}

The main paper reports the $\lambda$-sensitivity of Qwen-14B
on SQuAD~v2 and BoolQ across twelve values of $\lambda$. The
full numerical results, including the GSM8K math reasoning
dataset that was omitted from the main paper is reported in
Table~\ref{tab:lambda_extended}. We omit GSM8K from the main
paper's analysis because the absolute F1 on this benchmark falls
in the range $[0.005, 0.013]$ across all $\lambda$ values,
which is well within the noise floor of the evaluator under
our current pipeline. Reliable conclusions about
$\lambda$-sensitivity on math reasoning will require a
revised evaluation protocol with partial-credit scoring or a
chain-of-thought decoding setting, both of which lie outside
the scope of this paper.

\begin{table}[hbt!]
\centering
\small
\setlength{\tabcolsep}{4pt}
\caption{Extended $\lambda$-sweep validation F1 on Qwen-14B across SQuAD~v2, BoolQ, and GSM8K. Per-dataset optima are in bold.}
\label{tab:lambda_extended}
\begin{tabular}{lccc}
\toprule
$\lambda$ & SQuAD F1 & BoolQ F1 & GSM8K F1 \\
\midrule
0.000 & \textbf{0.8085} & 0.3597 & 0.0063 \\
0.001 & 0.7935 & 0.0534 & 0.0074 \\
0.003 & 0.8016 & 0.1461 & 0.0065 \\
0.010 & 0.4317 & 0.0710 & 0.0059 \\
0.030 & 0.3354 & 0.1068 & 0.0053 \\
0.070 & 0.3926 & 0.1733 & 0.0070 \\
0.100 & 0.2425 & \textbf{0.4218} & 0.0104 \\
0.200 & 0.2645 & 0.1905 & 0.0130 \\
0.350 & 0.3746 & 0.0996 & \textbf{0.0132} \\
0.500 & 0.1863 & 0.2001 & 0.0118 \\
0.750 & 0.3037 & 0.2960 & 0.0107 \\
1.000 & 0.1726 & 0.2076 & 0.0092 \\
\bottomrule
\end{tabular}
\end{table}

\subsection{$\mu$-Sensitivity Study}
\label{app:mu}

\begin{figure*}[t]
\centering
\includegraphics[width=\textwidth]{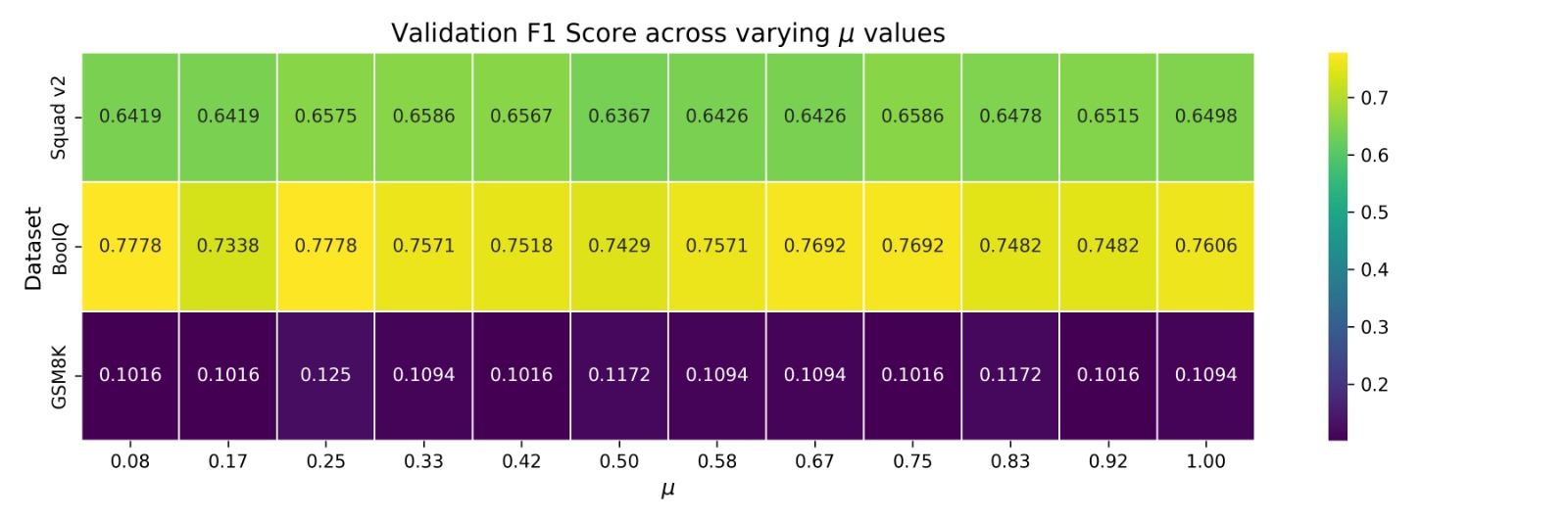}
\caption{Validation F1 across varying $\mu$ values for
Qwen-14B at $\lambda = 0.003$.}
\label{fig:mu_f1}
\end{figure*}

\begin{figure*}[t]
\centering
\includegraphics[width=\textwidth]{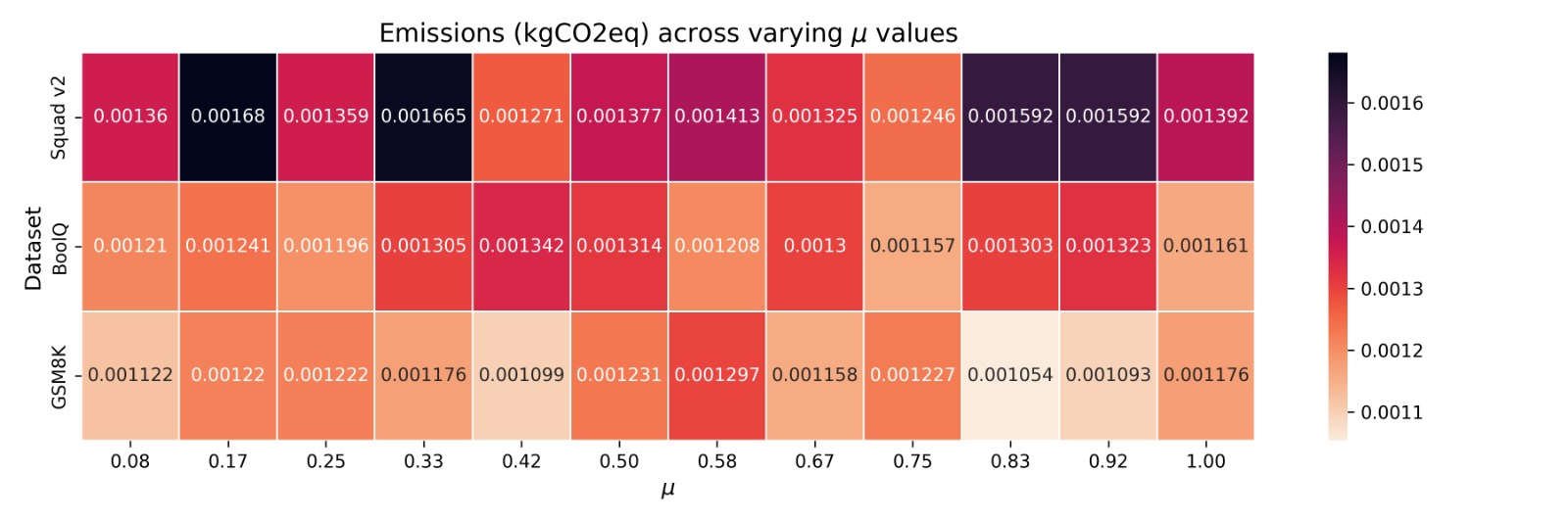}
\caption{Training emissions (kgCO$_2$eq) across varying $\mu$
values for Qwen-14B at $\lambda = 0.003$.}
\label{fig:mu_co2}
\end{figure*}

\begin{figure*}[hbt!]
\centering
\includegraphics[width=\textwidth]{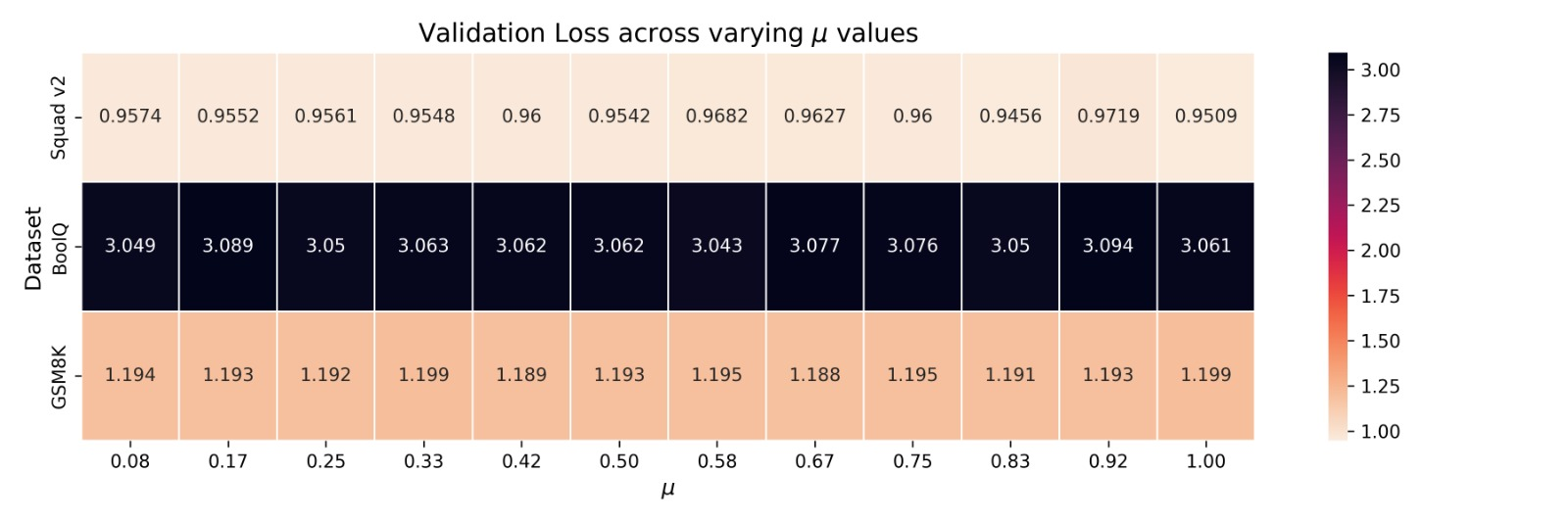}
\caption{Validation loss across varying $\mu$ values for
Qwen-14B at $\lambda = 0.003$.}
\label{fig:mu_loss}
\end{figure*}

In addition to the carbon-penalty weight $\lambda$, the
training objective admits an entropy-regularization term with
weight $\mu$ that we held fixed in the main paper for
exposition. Figures~\ref{fig:mu_f1}, \ref{fig:mu_co2}, and
\ref{fig:mu_loss} report the validation F1, emissions, and
validation loss for Qwen-14B across twelve values of $\mu$ on
the three sweep datasets at fixed $\lambda = 0.003$. The
F1 surface is essentially flat across $\mu$ for all three
datasets, with variation within $\pm 0.025$ F1 around the
mean; the emissions and loss surfaces are similarly flat. We
include these data for completeness and as evidence that the
$\mu$ term does not interact meaningfully with the carbon
penalty within the range studied.

\subsection{Per-Class AUC Analysis}
\label{app:per_class_auc}

The aggregate F1 numbers reported in Tables~\ref{tab:mmlu_main}
and \ref{tab:mmlu_full} can mask redistribution of accuracy
across answer classes. To examine whether the joint model's
gains are uniformly distributed, we report per-class AUC-PR
and AUC-ROC for the four MMLU answer options (A, B, C, D) on
the three main-paper models in Table~\ref{tab:per_class_auc}.
The reversals visible in Table~\ref{tab:qualitative} of the
main paper concentrate in the classes for which AUC improves
most under the joint objective, supporting the interpretation
that the carbon term is recovering specific decision
boundaries rather than diffusely shifting confidence across all
classes.

\begin{table}[hbt!]
\centering
\small
\setlength{\tabcolsep}{3pt}
\caption{Per-class AUC-PR for Qwen-14B on the three MMLU subjects, comparing CE baseline and joint ($\lambda = 0.1$). Largest per-row gain are in bold.}
\label{tab:per_class_auc}
\begin{tabular}{llcccc}
\toprule
Subject & Config & A & B & C & D \\
\midrule
\multirow{2}{*}{Abs.\ Algebra}
  & CE   & 0.561 & 0.691 & 0.601 & 0.489 \\
  & Joint & 0.591 & 0.696 & \textbf{0.628} & 0.552 \\
\multirow{2}{*}{Philosophy}
  & CE   & 0.793 & 0.875 & 0.918 & 0.823 \\
  & Joint & 0.799 & 0.870 & \textbf{0.926} & 0.814 \\
\multirow{2}{*}{Formal Logic}
  & CE   & 0.712 & 0.747 & 0.576 & 0.792 \\
  & Joint & 0.745 & 0.758 & \textbf{0.618} & 0.806 \\
\bottomrule
\end{tabular}
\end{table}

\subsection{Loss Curves}
\label{app:loss_curves}

\begin{figure*}[hbt!]
\centering
\includegraphics[width=\textwidth]{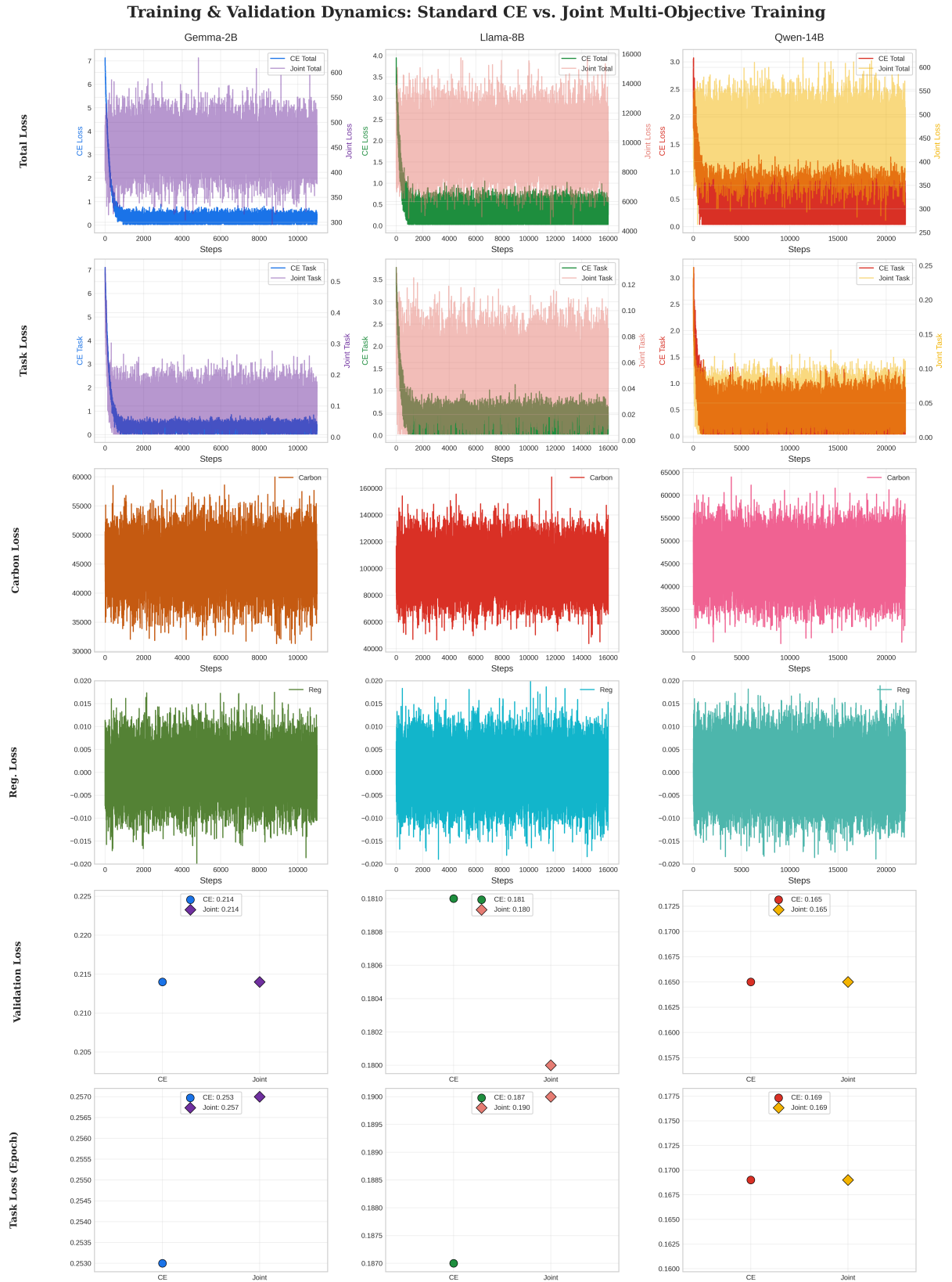}
\caption{Training-loss curves for Gemma-2B, Llama-8B, and
Qwen-14B, comparing the cross-entropy baseline and the joint
objective at the per-model selected $\lambda$.}
\label{fig:loss_curves}
\end{figure*}

Figure~\ref{fig:loss_curves} reports the training-loss curves
for the three models reported in the main paper, comparing the
cross-entropy baseline and the joint objective at the per-model
selected $\lambda$. The curves are visually similar in shape
and converge to similar terminal values, consistent with the
claim that the joint loss perturbs the optimum without
destabilizing the optimization itself.

\subsection{Extended Qualitative Examples}
\label{app:examples}

Table~\ref{tab:qualitative_full} extends the qualitative
examples of Table~\ref{tab:qualitative} in the main paper to
include the full item text, answer options, gold answer,
baseline prediction, and joint prediction for each example.
We provide additional examples from each of the three models
and three MMLU subjects to allow inspection of the kinds of
items the joint model recovers.

\begin{table*}[t]
\centering
\small
\renewcommand{\arraystretch}{1.25}
\setlength{\tabcolsep}{4pt}
\caption{Extended qualitative examples. Each row is
an item on which the cross-entropy baseline produced an incorrect answer and the joint model at $\lambda = 0.1$ produced the correct one. Probabilities are the per-option scores returned by
the model on the four answer choices.}

\label{tab:qualitative_full}
\begin{tabular}{p{0.10\linewidth} p{0.27\linewidth} p{0.27\linewidth} c c c}
\toprule
Subject & Question & Options (A / B / C / D) & Gold & CE $\to$ & Joint $\to$ \\
\midrule
Abstract \\ Algebra
 & Compute the product in the given ring. $(20)(-8)$ in $\mathbb{Z}_{26}$.
 & A. 0 \quad B. 1 \quad C. 11 \quad D. 22
 & D & A ($p{=}0.38$) & D ($p{=}0.43$) \\
\midrule
Abstract \\ Algebra
 & Statement 1: Every permutation is a cycle. Statement 2: Every cycle is a permutation.
 & A. True, True \quad B. False, False \quad C. True, False \quad D. False, True
 & D & A ($p{=}0.48$) & D ($p{=}0.73$) \\
\midrule
Formal Logic
 & Identify the antecedent of the conditional: \emph{``The Bees win their first game only if either the Aardvarks or the Chipmunks do not win their first games.''}
 & A. The Aardvarks do not win. \quad B. The Bees win their first game. \quad C. The Chipmunks do not win. \quad D. Either the Aardvarks or the Chipmunks do not win.
 & B & D ($p{=}0.52$) & B ($p{=}0.52$) \\
\midrule
Formal Logic
 & Construct a complete truth table for the argument $\sim\! C \supset D$; \; $D \supset C \,/\, C$. Then, using the truth table, determine whether the argument is valid or invalid.
 & A. Valid. \quad B. Invalid. Counterexample when $C$ and $D$ are true. \quad C. Invalid. Counterexample when $C$ is true and $D$ is false. \quad D. Invalid. Counterexample when $D$ is true and $C$ is false.
 & A & B ($p{=}0.31$) & A ($p{=}0.32$) \\
\midrule
Philosophy
 & According to Kant, the supreme principle of morality is:
 & A. analytic and a priori. \quad B. analytic and a posteriori. \quad C. synthetic and a priori. \quad D. synthetic and a posteriori.
 & C & A ($p{=}0.49$) & C ($p{=}0.57$) \\
\midrule
Philosophy
 & Aristotle says that what makes things be what they are --- their essence --- does not exist apart from individuals that exist in the world. If all the members of a species were destroyed, their essence or form:
 & A. would likewise be destroyed. \quad B. would be destroyed only if no one remembers the species. \quad C. would continue existing in some other realm of being. \quad D. would not be destroyed because there was no essence originally.
 & A & D ($p{=}0.59$) & A ($p{=}0.67$) \\
\bottomrule
\end{tabular}
\end{table*}

\subsection{Reproducibility Details}
\label{app:reproducibility}

All fine-tuning runs use AdamW with a constant learning rate,
batch sizes matched to each model's calibration set, and a
fixed number of optimization steps per (model, configuration)
pair. The fixed grid carbon intensity is
$0.369473~\mathrm{kg\,CO_2/kWh}$, corresponding to the
training region; all relative CO$_2$ comparisons in the paper
are invariant to this constant. Energy is logged with
CodeCarbon at per-step granularity. We release the per-step
training histories, calibration tables, inference-time
emission logs, MMLU prediction files for all configurations,
and the surrogate-weight JSON files referenced in
Tables~\ref{tab:calibration} and \ref{tab:surrogate_weights}.
The release supports independent recomputation of every
numerical claim in the main paper and in the Appendix.

\subsection{Discussion}
\label{sec:discussion}

\paragraph{The non-empty but selective break-even region.}
Our central empirical finding is that joint optimization of task
performance and inference carbon admits operating points at which
F1 is preserved or improved while inference CO$_2$ is held at or
below the cross-entropy baseline. The strongest of these
operating points, Qwen-2.5-14B on abstract algebra, with a
$3.5$-point F1 gain and a $3.5\%$ CO$_2$ reduction, is a strict
Pareto improvement, and two further pairs (Gemma-2-2B on
philosophy and Llama-3.1-8B on formal logic) deliver substantial
F1 gains at essentially zero or low carbon cost. The result we
do \emph{not} report is equally important: the break-even
region is not universal. Four of nine (model, subject) pairs in
Table~\ref{tab:mmlu_main} fall outside it, and the
$\lambda$-sensitivity study in Figure~\ref{fig:lambda_sensitivity}
shows that SQuAD~v2 strictly degrades under any nonzero
$\lambda$. We read these two facts together as supporting a
specific interpretation: $\lambda$ acts as a structural
regularizer whose sign of effect is determined by the geometry
of the target task, not as a uniform efficiency intervention.
Tasks that admit a lower-cost computational pathway compatible
with the correct answer (boolean reasoning, multi-step symbolic
manipulation) benefit from a mild implicit-complexity prior;
tasks that require exact span selection (extractive QA) are
brittle to any objective perturbation. Practitioners should
expect the optimal $\lambda$ to be calibrated per task rather
than transferred across them.

\paragraph{Concentrated effect at inference time.}
Training-side emissions across $\lambda$ values are within
measurement noise (Table ~\ref{tab:lambda_sweep_full}), which
might at first appear to undercut the framing of a
``carbon-aware'' objective. We argue the opposite: it clarifies
where the mechanism actually operates. The joint loss does not
reduce the cost of producing the model. Fine-tuning a
fixed-architecture LLM for a fixed number of steps incurs
approximately the same energy regardless of which scalar penalty
is added to the cross-entropy term. What the joint loss does
instead is shift the optimum to which fine-tuning converges,
selecting parameter configurations whose forward pass the
surrogate predicts to be cheaper to execute. The carbon-aware
character of the resulting model is therefore a property of the
deployed checkpoint rather than of the fine-tuning run that
produced it, and the empirical signature of the mechanism
appears on the inference-time emission columns of
Table~\ref{tab:mmlu_main} rather than on the training-time
columns. This positioning aligns the contribution with the
lifecycle observation that motivated the paper: that the
operational carbon of a widely deployed LLM is dominated by its
serving footprint rather than its training footprint
\citep{patterson2021carbon, wu2022sustainable}.




\end{document}